\documentclass[conference, hidelinks]{IEEEtran}
\usepackage{times}

\usepackage[numbers]{natbib}
\usepackage{multicol}

\usepackage{mathtools}
\usepackage[pdftex,final]{graphicx}
\usepackage{units}
\usepackage[english]{babel}
\usepackage[latin1]{inputenc}
\usepackage[usenames]{color}
\usepackage{comment}
\usepackage{balance}
\usepackage{url}
\usepackage{todonotes}
\usepackage{amssymb}
\usepackage{tikz}
\usetikzlibrary{calc}
\usetikzlibrary{mindmap,trees}

\usepackage{booktabs}

\usepackage{xcolor}

\definecolor{darkblueish}{RGB}{64, 87, 116}
\definecolor{blueish}{RGB}{103, 135, 176}
\definecolor{greenish}{RGB}{177, 177, 123}
\definecolor{reddish}{RGB}{ 205, 102, 7}
\definecolor{orangeish}{RGB}{246, 160, 61}
\colorlet{bluegreenish}{blueish!50!greenish}

\usepackage[bookmarks=true]{hyperref}
\hypersetup{
    colorlinks,
    linkcolor={reddish!70!black},
    citecolor={blueish!70!black},
    urlcolor={blueish!70!black}
}
\usepackage[all]{hypcap}

\usepackage{algorithm}
\usepackage{algpseudocode}
\usepackage{setspace}
\usetikzlibrary{bayesnet}

\usepackage{todonotes}

\newcommand{\sta}{x} 
\newcommand{\obs}{y} 

\newcommand{\pre}{1}
\newcommand{\cur}{2}
\newcommand{\pos}{3}

\usepackage{flushend}

\makeatletter
\def\input@path{{../figures/}}
\makeatother
\graphicspath{{../figures/}}

\begin{document}

\title{A New Perspective and Extension\\ of the Gaussian Filter}


%
%
%

%
%
%

%
\author{\authorblockN{Manuel W\"uthrich\authorrefmark{1},
Sebastian Trimpe\authorrefmark{1},
Daniel Kappler\authorrefmark{1} and
Stefan Schaal\authorrefmark{1}\authorrefmark{2}}

\authorblockA{\authorrefmark{1}Autonomous Motion Department\\
Max Planck Institute for Intelligent Systems,
T\"ubingen, Germany\\
Email: first.lastname@tuebingen.mpg.de}

\authorblockA{\authorrefmark{2}Computational Learning and Motor Control lab\\
University of Southern California,
Los Angeles, CA, USA}
}

\maketitle

\begin{abstract}
The Gaussian Filter (GF) is one of the most widely used filtering algorithms; instances are the Extended Kalman Filter, the Unscented Kalman Filter and the Divided Difference Filter. GFs represent the belief of the current state by a Gaussian with the mean being an affine function of the measurement. We show that this representation can be too restrictive to accurately capture the dependences in systems with nonlinear observation models, and we investigate how the GF can be generalized to alleviate this problem.  To this end, we view the GF from a variational-inference perspective. We analyse how restrictions on the form of the belief can be relaxed while maintaining simplicity and efficiency. This analysis provides a basis for generalizations of the GF. We propose one such generalization which coincides with a GF using a virtual measurement, obtained by applying a nonlinear function to the actual measurement. Numerical experiments show that the proposed Feature Gaussian Filter (FGF) can have a substantial 
performance advantage over the standard GF for systems with nonlinear observation models. 
\end{abstract}

\IEEEpeerreviewmaketitle


\section{Introduction}\label{sect:introduction}
Decision making requires knowledge of some variables of interest.
In the vast majority of real-world problems, these variables are latent,
i.e. they cannot be observed directly and must be inferred from available measurements. 
To maintain an up-to-date belief over the latent
variables, past measurements have to be fused continuously with incoming
measurements.
This process is called filtering and its applications range from robotics
to estimating a communication signal using noisy measurements.

\subsection{Dynamical Systems Modelling}
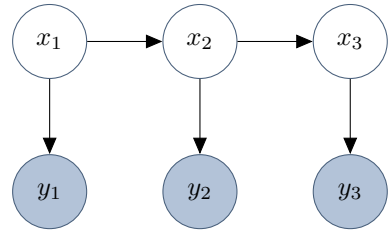
\begin{figure}[tb]	
  \begin{center}
      \begin{tikzpicture}
   
   \node[latent, draw=blueish!70!black]                      	  	(sta_pre) {$\sta_{\pre}$};
   \node[latent, right=of sta_pre, draw=blueish!70!black]          	(sta_cur) {$\sta_{\cur}$};
   \node[latent, right=of sta_cur, draw=blueish!70!black]          	(sta_pos) {$\sta_{\pos}$};
    
   \node[obs, below=of sta_pre, fill=blueish!50!white, draw=blueish!70!black]                  		(obs_pre) {$\obs_{\pre}$};
   \node[obs, below=of sta_cur, fill=blueish!50!white, draw=blueish!70!black]         	(obs_cur) {$\obs_{\cur}$};
   \node[obs, below=of sta_pos, fill=blueish!50!white, draw=blueish!70!black]         	(obs_pos) {$\obs_{\pos}$};
    
  \edge {sta_pre} {sta_cur} ; %
  \edge {sta_cur} {sta_pos} ; %
  
  \edge {sta_pre} {obs_pre}
  \edge {sta_cur} {obs_cur}
  \edge {sta_pos} {obs_pos}

\end{tikzpicture} 
  \end{center}
  \caption{The belief network which characterizes the evolution of 
  the state $\sta$ and the observations $\obs$.}
  \label{fig:standard_system}
\end{figure}

Dynamical systems are typically modelled in a state-space
representation, which means that the state is chosen such that the
following two statements hold.
First, 
the current observation depends only on the current state.
Secondly, the next state of the system depends only on the current
state.
These assumptions can be visualized by the belief network shown in
Figure \ref{fig:standard_system}.

We assume the system to be stationary, i.e. there is no explicit
dependence on time.
Therefore, the absolute time indices are irrelevant. Only the time
difference within a figure or equation is of importance.
To simplify notation, we will use the indices $1,2,3$ throughout the
paper.

A stationary system can be characterized by two functions.
The process model
\begin{align}
x_{ 2 }=g(x_{ 1 },v_{ 2 })\label{eq:functional_process_model}
\end{align}
describes the evolution of the state.
Without loss of generality, we can assume the noise $v_2$ to be drawn from a
Gaussian with zero mean and unit variance, since it can always be mapped onto any other distribution
inside of the nonlinear function $g(\cdot)$.
The observation model
\begin{align}
 y_{ 2 }=h(x_{ 2 },w_{ 2 })\label{eq:functional_observation_model}
\end{align}
describes how a measurement is produced from the current state.
Following the same reasoning as above, we assume the noise $w_2$ to be Gaussian with zero mean and unit variance.
The process and observation models can also be represented by
distributions.
The distributional form of both models are implied by their functional
form
\begin{align}
 p(x_{ 2 }|x_{ 1 }) &=\int\limits_{ v_{ 2 } } \delta (x_{ 2 }-g(x_{ 1 },v_{ 2 }))p(v_{ 2 })\label{eq:distributional_process_model}\\ 
 p(y_{ 2 }|x_{ 2 }) &=\int\limits_{ w_{ 2 } } \delta (y_{ 2 }-h(x_{ 2 },w_{ 2 }))p(w_{ 2 })\label{eq:distributional_observation_model}
\end{align}
where $\delta$ is the Dirac delta function.
While both representations contain the exact same information,
sometimes one is more convenient than the other.

\subsection{Exact Filtering}
The desired posterior distribution over the current state $p(x_{ 2 }|y_{ :2 })$ can be computed recursively from the distribution over the previous state $p(x_{ 1 }|y_{ :1 })$; the subscript $({:t})$ denotes all time steps up to $t$. This recursion can be written in two steps, a prediction step
\begin{align}
 p(x_{ 2 }|y_{ :1 })=\int\limits_{ x_{ 1 } } p(x_{ 2 }|x_{ 1 })p(x_{ 1 }|y_{ :1 })\label{eq:prediction}
\end{align}
and an update step
\begin{align}
 p(x_{ 2 }|y_{ :2 })=\frac { p(y_{ 2 }|x_{ 2 })p(x_{ 2 }|y_{ :1 }) }{ \int\limits_{ x_{ 2 } } p(y_{ 2 }|x_{ 2 })p(x_{ 2 }|y_{ :1 }) } .\label{eq:filtering}
\end{align}
\citet{kalman1960new} found the solution to these equations for linear process and observation models with additive Gaussian noise. However, filtering in nonlinear systems remains an important area of research. Exact solutions \cite{benes, daum} have been found for only a very restricted class of process and observation models. For more general dynamical systems, it is well known that the exact posterior distribution cannot be represented by a finite number of parameters \cite{earlyKushner}. Therefore, the need for approximations is evident. 

\subsection{Approximate Filtering}
Approximate filtering methods are typically divided into
deterministic, parametric methods, such as the Unscented Kalman Filter
(UKF) \cite{ukf} and the Extended Kalman Filter (EKF) \cite{ekf}, and
stochastic, nonparametric methods such as the Particle Filter (PF) 
\cite{gordon}.
In this paper, we argue that there is a more fundamental division between filtering methods.

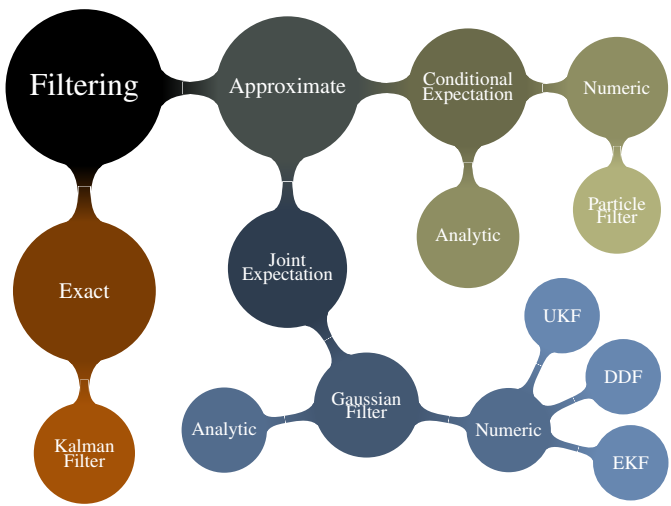
\begin{figure}[bt]	
  \begin{center}
      \newcommand{\scaleOne}{1.3}
\newcommand{\distanceOne}{10.0cm}
\newcommand{\distanceTwo}{3.5cm}
\newcommand{\distanceThree}{3.0cm}
\newcommand{\distanceFour}{2.7cm}
\newcommand{\widthOne}{2cm}
\newcommand{\sizeOne}{3.3cm}
\newcommand{\sizeTwo}{3.0cm}
\newcommand{\sizeThree}{2.4cm}
\newcommand{\sizeFour}{2.1cm}
\newcommand{\sizeFive}{1.7cm}
\newcommand{\sizeSix}{1.5cm}

\tikzset{every node/.append style={scale=1.0}}

\begin{tikzpicture}[scale=0.6,transform shape]

  \path[mindmap,concept color=black,text=white]
    node[concept , minimum size=\sizeOne, text width=\sizeOne,] {\huge Filtering} 
    [clockwise from=0]
    child[concept color=reddish!60!black, grow=-90, level distance=4.5cm] 
    {
      node[concept , minimum size=\sizeTwo, text width=\sizeTwo] {\Large Exact}
      [clockwise from=0]
      child[concept color=reddish!80!black, grow=-90, level distance=3.6cm] {
        node[concept , minimum size=2cm, text width=2cm] {\large Kalman Filter}
      }
    }  
    child[concept color=bluegreenish!50!black, grow=0, level distance=4.5cm] 
    {
      node[concept , minimum size=\sizeTwo, text width=\sizeTwo ] {\Large Approximate}
      [clockwise from=0]
      child[concept color=greenish!60!black, grow=0, level distance=4cm] {
        node[concept , minimum size=\sizeThree, text width=\sizeThree ] {\large Conditional Expectation} 
        [clockwise from=0]
        child[concept color=greenish!80!black, grow=-90, level distance=3.3cm] {
          node[concept , minimum size=\sizeFour, text width=\sizeFour] {\large Analytic} 
        }
        child[concept color=greenish!80!black, grow=0 , level distance=3.3cm]{
          node[concept , minimum size=\sizeFour, text width=\sizeFour] {\large Numeric} 
          [clockwise from=0]
          child[concept color=greenish!100!black, grow=-90, level distance=\distanceFour]{
            node[concept , minimum size=\sizeFive, text width=\sizeFive] {\large Particle Filter}
          }
        }
      }
      child[concept color=blueish!45!black, grow=-90, level distance=4cm ]{
        node[concept , minimum size=\sizeThree, text width=\sizeThree] {\large Joint Expectation} 
        [clockwise from=0]
        child[concept color=blueish!65!black, grow=-60,level distance=3.5cm] {
          node[concept , minimum size=\sizeFour, text width=\sizeFour] {\large Gaussian Filter} 
          [clockwise from=0]
          child[concept color=blueish!80!black, grow=-10, level distance=3.2cm] {
            node[concept, minimum size=\sizeFive, text width=\sizeFive] {\large Numeric} 
            child[concept color=blueish!100!black, grow=65, level distance=2.8cm]{
              node[concept , minimum size=\sizeSix, text width=\sizeSix] {\large UKF}
            }
            child[concept color=blueish!100!black, grow=25 , level distance=2.8cm]{
              node[concept , minimum size=\sizeSix, text width=\sizeSix] {\large DDF}
            }
            child[concept color=blueish!100!black, grow=-15 , level distance=2.8cm]{
              node[concept , minimum size=\sizeSix, text width=\sizeSix] {\large EKF}
            }
          }
          child[concept color=blueish!80!black, grow=190, level distance=3.2cm] {
            node[concept, minimum size=\sizeFive, text width=\sizeFive] {\large Analytic} 
          }
        }
      }
    };
\end{tikzpicture} 
  \end{center}
  \caption{A taxonomy of filtering algorithms.}
  \label{fig:taxonomy}
\end{figure}

To the best of our knowledge, all existing filtering algorithms either
compute expectations with respect to the conditional distribution
$p(x_2|y_{:2})$ or with respect to the joint distribution $p(x_2, y_2|y_{:1})$.
In Figure \ref{fig:taxonomy}, we divide approximate filtering
algorithms according this criterion.
The computational power required to numerically compute expectations with respect to $p(x_2|y_{:2})$
increases exponentially with the state dimension, limiting the use of such methods to low dimensional problems. In contrast, expectations
with respect to the joint distribution $p(x_2, y_2|y_{:1})$ can be approximated numerically with linear complexity in the state dimension.
In Section \ref{sect:approx_update}, we show how this fundamental difference arises.

Since conditional expectation methods suffer from the curse of dimensionality, we focus on joint
expectation methods in this paper.
To the best of our knowledge, all such methods approximate the true 
joint distribution $p(x_2, y_2|y_{:1})$ with a Gaussian distribution
$q(x_2, y_2|y_{:1})$ and subsequently condition on $y_2$, which is
easy due to the Gaussian form. This approach is called the Gaussian
Filter, of which the well known EKF \cite{ekf}, the
UKF \cite{ukf} and the Divided Difference
Filter (DDF) \cite{ddf} are instances \cite{wu,gf}.

\citet{garcia} show that for nonlinear dynamical systems, Gaussians can yield a
poor fit to the true joint distribution $p(x_2,y_2|y_{:1})$, which 
in turn leads to bad filtering performance. To address this problem, we search for 
a more flexible representation of the belief that can accurately capture the 
dependences in the dynamical system, while maintaining the efficiency of the GF.

In Sections \ref{sect:approx_pred} to \ref{sect:gaussian_filter}, we first review existing filtering methods, in particular the GF.
Then we find some desiderata for the form 
of the approximate belief in Section \ref{sect:generalization} to provide a basis for efficient generalizations of the GF.
In Section \ref{sect:feature_gaussian_filter}, we propose one possible form of 
the approximate belief 
and show that this generalization coincides with the GF using a virtual measurement given by a nonlinear function of the actual measurement.
Numerical examples in Section \ref{sect:PerformanceAnalysis} highlight the potential performance gains of the proposed filter over the standard GF.

\section{Approximate Prediction}
\label{sect:approx_pred}
We start out with the distribution $p(x_{ 1 }|y_{ :1 })$ computed in the 
previous time step. 
The representation of the beliefs might be
parametric, such as a Gaussian, or it might be nonparametric, e.g.
represented by a set of samples. In any case, the goal is to find the
prediction $p(x_{ 2 }|y_{ :1 })$ given the previous belief. When there
is no closed form solution to (\ref{eq:prediction}), we have to
settle for finding certain properties of the predicted belief
$p(x_{ 2 }|y_{ :1 })$ instead of the full distribution. 
For all filtering algorithms we are aware of, these desired properties can be
written as expectations
\begin{align}
 \int\limits_{ x_{ 2 } } f(x_{ 2 })p(x_{ 2 }|y_{ :1 }).\label{eq:predicted_expectation}
\end{align}
For instance with $f(x_2)=x_2$, we obtain the mean $\mu$, and with
$f(x_2)=(x_2-\mu)(x_2-\mu)^T$, we obtain the covariance. These expectations can then be used to
find the parameters of an approximate distribution. A widely used approach is
moment matching,  where the moments of the
approximate distribution are set to the moments of the
exact distribution. We will analyse such methods in more detail below. 
What is important here is that we are always concerned with finding
expectations of the form of (\ref{eq:predicted_expectation}).

We substitute (\ref{eq:prediction}) in (\ref{eq:predicted_expectation}) in order to write this expectation in terms of the last belief and the process model:
\begin{align}
 \int\limits_{ x_{ 2 } } f(x_{ 2 })p(x_{ 2 }|y_{ :1 })=\int\limits_{ x_{ 2 } } f(x_{ 2 })\int\limits_{ x_{ 1 } } p(x_{ 2 }|x_{ 1 })p(x_{ 1 }|y_{ :1 }) .
\end{align}
Substituting the distributional process model (\ref{eq:distributional_process_model}) and solving the integral over $x_2$, which is easy due to the Dirac distribution $\delta$, we obtain
\begin{align}
\boxed{
 \int\limits_{ x_{ 2 } } f(x_{ 2 })p(x_{ 2 }|y_{ :1 })=\int\limits_{ x_{ 1 },v_{ 2 } } f(g(x_{ 1 },v_{ 2 }))p(v_{ 2 })p(x_{ 1 }|y_{ :1 }) .
 }\label{eq:predicted_expectation_full}
\end{align}
For certain process models $g$ and functions $f$, it is possible to find a closed form
solution. In general, however, this integral has to be computed
numerically. Since $p(v_2)$ is the Gaussian noise distribution and
$p(x_{ 1 }|y_{ :1 })$ is the previous belief in the
representation of choice, it is generally possible to sample from these two
distributions. This is crucial since it allows for efficient numerical
integration.

One possibility is to use Monte Carlo sampling to approximate the expectation from (\ref{eq:predicted_expectation_full}). The standard deviation of the estimate is proportional to $\frac{1}{\sqrt{L}}$, with $L$ being the number of samples. The dimension of the state does not affect the standard deviation of the estimate \cite{owen}.

Another possibility is to use deterministic numerical integration algorithms, such as Gaussian quadrature methods. The complexity of such methods typically scales linearly with the state dimension \cite{wu}.

Which particular numeric integration method is used to compute the approximate expectations is
inconsequential for the results presented in this paper. What is important is that expectations of the type required in the prediction step can
be approximated efficiently, even for a high dimensional state. This is
unfortunately not the case for the update step, which is the issue
we are addressing in this paper.

\section{Approximate Update}
\label{sect:approx_update}
The goal of the update step is to obtain an approximation of the posterior $p(x_2|y_{:2})$, based on the belief $p(x_{ 2 }|y_{ :1 })$ which has been computed in the prediction step. 

\subsection{Computation of Conditional Expectations}
As for the prediction, when there is no exact solution to (\ref{eq:filtering}), we compute expectations with respect to the posterior $\int\nolimits_{ x_{ 2 } } r(x_{ 2 })p(x_{ 2 }|y_{ :2 })$, where $r(\cdot)$ is an arbitrary function. 
We insert (\ref{eq:filtering}) to express this expectation in terms of the observation model and the predicted distribution:
\begin{align}
 \int\limits_{ x_{ 2 } } r(x_{ 2 })p(x_{ 2 }|y_{ :2 })=\frac { \int\limits_{ x_{ 2 } } r(x_{ 2 })p(y_{ 2 }|x_{ 2 })p(x_{ 2 }|y_{ :1 }) }{ \int\limits_{ x_{ 2 } } p(y_{ 2 }|x_{ 2 })p(x_{ 2 }|y_{ :1 }) }  .
\end{align}
Both the numerator and the denominator can be written as 
\begin{align}
\boxed{
 \int\limits_{ x_{ 2 } } f(x_{ 2 })p(y_{ 2 }|x_{ 2 })p(x_{ 2 }|y_{ :1 })
 }\label{eq:filtered_expectation_2}
\end{align}
with $f(x)=r(x)$ for the numerator and $f(x)=1$ for the denominator. The update step thus amounts to computing expectations of the form of (\ref{eq:filtered_expectation_2}).

As in the prediction step, we can approximate this expectation either by
sampling, which is used in Sequential Monte Carlo (SMC) \cite{gordon,
  cappe_overview}, or by applying deterministic methods such as
Gaussian quadrature \cite{kushner}.

There is, however, a very important difference to the prediction step. We now need to compute the expectation of a function $f$ weighted with the observation model $p(y_{ 2 }|x_{ 2 })$. If these weights are very small at most evaluation points, the numeric integration becomes inaccurate, an effect known as particle deprivation in particle filters \cite{cappe_overview}.

Unfortunately, this effect becomes worse with increasing dimensionality. To see this, consider a simple example with predictive distribution $p(x_{ 2 }|y_{ :1 })=\mathcal{N}(x_2|0,I)$ and observation model $p(y_{ 2 }|x_{ 2 })=\mathcal{N}(y_{ 2 }|x_{ 2 },I)$. Both the state and measurement dimensions are equal to $D$. Computing the expected weight, i.e. the expected value of the likelihood, yields
\begin{align}
 E[p(y_2|x_2)]\!=\!\!\!\!\int\limits_{ x_2,y_2 }\!\!\!\!\! p(y_2|x_2)p(y_2|x_2)p(x_2|y_{:1})=(2 \sqrt{\pi})^{-D}\!\!.\!\!
\end{align}
That is, the expected weight decreases exponentially with the dimension $D$. In fact, it is well known that the computational demands of such methods increase exponentially with the state dimensionality \cite{li, bickel, owen}. Thus, methods that rely on the computation of conditional expectations are restricted to dynamical
systems which either have a simple structure such that expectations can
be computed analytically, or are low dimensional such that
numeric methods can be used. 

\subsection{Computation of Joint Expectations}
There are a number of approaches which avoid computing such expectations with respect to the conditional distribution $p(x_{ 2 }|y_{ :2 })$. Instead, these methods express the parameters of the approximate posterior $q(x_{ 2 }|y_{ :2 })$ as a function of expectations with respect to the joint distribution:
\begin{align}
\int\limits_{ x_{ 2 },y_{ 2 } } \!\!\!\!\! f(x_{ 2 },y_{ 2 })p(y_{ 2 },x_{ 2 }|y_{ :1 })\!=\!\!\!\!\!\int\limits_{ x_{ 2 },y_{ 2 } } \!\!\!\!\! f(x_{ 2 },y_{ 2 })p(y_{ 2 }|x_{ 2 })p(x_{ 2 }|y_{ :1 }) \!\! \label{eq:joint_expectation}
\end{align}
Inserting the observation model from (\ref{eq:distributional_observation_model}) into the joint expectation above and solving the integral over $y_2$ yields
\begin{equation}
   \boxed{
   \begin{aligned}
\int\limits_{ x_{ 2 },y_{ 2 } } &f(x_{ 2 },y_{ 2 })p(y_{ 2 },x_{ 2 }|y_{ :1 })=\\
&\int\limits_{ x_{ 2 },w_{ 2 } } f(x_{ 2 },h(x_{ 2 },w_{ 2 }))p(w_{ 2 })p(x_{ 2 }|y_{ :1 }) .
   \end{aligned}
   }\label{eq:objective_final}
\end{equation}
This term has the same form as the expectation in the prediction step (\ref{eq:predicted_expectation_full}). It is an integral of an arbitrary function with respect to probability densities that can be sampled.
This allows us to approximate this expectation efficiently, even for high dimensional states.

\subsection{Conclusion}
The insight of this section is that computing expectations numerically with
respect to the conditional distribution $p(x_2|y_{:2})$ requires
exponential computational power in the state dimension, whereas the
complexity of computing expectations with respect to the joint
distribution $p(x_2,y_2|y_{:1})$ scales linearly with the state
dimension. Note that expectations with respect to the
marginals $p(x_2|y_{:1})$ and $p(y_2|y_{:1})$ are a special case of an
expectation with respect to the joint distribution and can be
computed efficiently as well.

In the remainder of the paper, we only consider the update step. 
Thus, the only variables we require are $x_2$ and $y_2$; $x_1$ will
not be considered. Therefore, we drop
the indices for ease of notation. Furthermore, we 
make the dependence on $y_{:1}$ implicit.  That is, $p(x_2,y_2|y_{:1})$ becomes
$p(x,y)$ and $p(x_2|y_{:2})$ becomes $p(x|y)$, etc.

\section{The Gaussian Filter}\label{sect:gaussian_filter}
The advantage in terms of computational complexity of joint
expectation filters over conditional expectation filters comes
at a price: The approximate posterior $q(x|y)$ must have a
functional form such that its parameters can be computed efficiently
from these joint expectations. To the best of our knowledge, all
existing joint expectation filters solve this issue by approximating
the true joint distribution $p(x,y)$ with a Gaussian distribution:
\begin{align}
q(x,y)=\mathcal{N}\left(\begin{pmatrix} x \\ y \end{pmatrix} \Big|  \begin{pmatrix} \mu _{ x } \\ \mu _{ y } \end{pmatrix},\begin{pmatrix} \Sigma _{ xx } & \Sigma _{ xy } \\ \Sigma _{ yx } & \Sigma _{ yy } \end{pmatrix} \right).
\end{align}
The parameters of this approximation are readily obtained by moment matching, i.e. the moments of the Gaussian are set to the moments of the exact distribution:
\begin{equation}
   \boxed{
   \begin{aligned}
\mu _{ x }&=\int\limits_{ x } xp(x)\\ 
\mu _{ y }&=\int\limits_{ y } yp(y)\\ 
\Sigma _{ xx }&=\int\limits_{ x } (x-\mu _{ x })(x-\mu _{ x })^{ T }p(x)\\ 
\Sigma _{ yy }&=\int\limits_{ y } (y-\mu _{ y })(y-\mu _{ y })^{ T }p(y)\\ 
\Sigma _{ xy }&=\int\limits_{ x,y } (x-\mu _{ x })(y-\mu _{ y })^{ T }p(x,y).
   \end{aligned}
   }\label{eq:gf_expectations}
\end{equation}
All of these expectations can be computed efficiently for reasons explained in the previous section.

 After the moment matching step, we condition on $y$ to obtain the desired posterior, which is a simple operation since the approximation is Gaussian:
\begin{align}
\hspace{-0.2cm} q(x&|y)\!=\!\mathcal{N}\!(x|\mu _{ x }\!+\!\Sigma _{ xy }{ \Sigma  }_{ yy }^{ -1 }(y\!-\!\mu _{ y }),\Sigma _{ xx }\!-\!\Sigma _{ xy }{ \Sigma  }_{ yy }^{ -1 }{ \Sigma  }_{ xy }^{ T }).\! \label{eq:gaussian_filter_posterior}
\end{align}
This approach is called the Gaussian Filter (GF) \cite{gf, sarkka, wu}. Widely used filters such as the EKF \cite{ekf}, the UKF \cite{ukf} and the DDF \cite{ddf} are instances of the Gaussian Filter, differing only in the numeric integration method used for computing the expectations in (\ref{eq:gf_expectations}).

While much effort has been devoted to finding accurate numeric integration schemes for computing these expectations, there seems to be no joint expectation method using a non-Gaussian joint approximation $q(x,y)$. The posterior \eqref{eq:gaussian_filter_posterior}, which we ultimately care about, is therefore Gaussian in the state $x$ with the mean being an affine function of $y$. As we show in the experimental section, this form can be too restrictive to accurately capture the relationship between the measurement and the state in nonlinear settings. This leads to information about the state being discarded and ultimately to poor filtering performance.

\section{Generalization of the Gaussian Filter}\label{sect:generalization}
In this section, we investigate whether it is possible to find a more general form of the approximate posterior $q(x|y)$ that still allows for efficient computation of the parameters.
To this end, we write the problem of finding the parameters of the approximation as an optimization problem.

In the GF, the parameters $\Theta$ of the Gaussian belief $q(x,y|\Theta)$ are found by moment matching. For a Gaussian approximation, moment matching is equivalent to minimizing the KL-divergence \cite{barber}
\begin{align}
 \mathrm{KL}[p(x,y)|q(x,y|\Theta )]=\int\limits_{ x,y } \log  \left(\frac { p(x,y) }{ q(x,y|\Theta ) } \right)p(x,y).\label{eq:kl}
\end{align}
By minimizing \eqref{eq:kl} with respect to $\Theta$, we can thus retrieve the GF. Furthermore, the KL-divergence has convenient analytic properties. It is a widely used objective for matching distributions and  can be justified from an information theoretic point of view \cite{barber}.

Having found an appropriate objective for the approximation, it is natural to ask if it is possible to find more general, non-Gaussian approximations. The form of $q(x,y|\Theta)$ is restricted by the requirement of being able to condition on $y$ in closed form in order to find the approximate conditional $q(x|y,\Theta)$. This requirement is met automatically if we choose the form of the conditional distribution and the marginal distribution separately, instead of picking a form for the joint distribution. The joint distribution is then given by $q(x,y|\Theta)=q(x|y,\theta)q(y|\vartheta)$, where we have split the parameter set $\Theta$ into $\theta$ and $\vartheta$.
Any conditional and marginal distributions can be combined to form a valid joint distribution. Hence, the respective parameter sets $\theta$ and $\vartheta$ can be chosen independently. Imposing any constraints tying the two parameter sets together would restrict the flexibility of the joint distribution unnecessarily. 

Inserting this factorization into (\ref{eq:kl}), we obtain
\begin{align}
\mathrm{KL}[p(x,y)|q(x,y|\Theta )]\!=\!c(\vartheta ) + \mathrm{KL}[p(x,y)|q(x|y,\theta )]
\end{align}
where we have collected all terms independent of $\theta$ in $c(\vartheta)$. Since only the conditional distribution is of interest, we will maximize with respect to $\theta$. Hence, we can drop the terms which do not depend on $\theta$, which leads to the objective function
\begin{align}
\boxed{
\mathrm{KL}[p(x,y)|q(x|y,\theta )]=\int\limits_{ y,x } \log  \left(\frac{p(x,y)}{q(x|y,\theta )}\right)p(x,y).
}\label{eq:objective}
\end{align}
Note that this is a somewhat unusual KL-divergence, since it compares a joint distribution with a conditional distribution. However, this configuration is very desirable in this context. We can directly obtain the approximate posterior distribution $q(x|y,\theta)$ from the exact joint distribution $p(x,y)$ by minimizing \eqref{eq:objective} with respect to $\theta$. Only expectations with respect to the joint distribution $p(x,y)$ are required, and we have seen that these can be approximated efficiently.

\subsection{Desiderata for the Form of the Approximation}
In the following, we seek conditions on the form of $q(x|y,\theta)$ that allow for an efficient minimization of (\ref{eq:objective}) with respect to $\theta$.

First, $q(x|y,\theta )$ 
has to integrate to one in $x$ since it is a probability distribution. We can enforce this condition by writing 
\begin{align}
q(x|y,\theta )=\frac { r(x,y,\theta ) }{ \int_x r(x,y,\theta ) } 
\end{align}
with $r(x,y,\theta )$ being any positive function whose integral in $x$ over the real domain is finite and non-zero.

Furthermore, for the objective in (\ref{eq:objective}) to be well defined, the support of $q(x|y,\theta )$ has to contain the support of $p(x,y)$. Since $p(x,y)$ could be any distribution, we will choose the form $q(x|y,\theta )$ such that it has infinite support; that is, $q(x|y,\theta )>0$ everywhere, which implies $r(x,y,\theta )>0$. This condition is enforced by writing the approximate distribution as
\begin{align}
\boxed{
q(x|y,\theta )=\frac { { e }^{ f(x,y,\theta ) } }{ \int_x { e }^{ f(x,y,\theta ) } }
}\label{eq:approximate_distribution}
\end{align}
with $f(x,y,\theta )=\log(r(x,y,\theta ))$. The question we will address in the following is what $f$ has to look like in order to obtain an efficient filtering algorithm.

Substituting $q(x|y,\theta )$ in (\ref{eq:objective}), we obtain
\begin{equation}
   \boxed{
   \begin{aligned}
&\mathrm{KL}[p(x,y)|q(x|y,\theta )]=C+\\
&\int\limits_{ y } \log  \left(\int\limits_{ x }{ e } ^{ f(x,y,\theta ) }\right)p(y)-\int\limits_{ y,x } f(x,y,\theta )p(x,y)
   \end{aligned}
   }\label{eq:objective_final}
\end{equation}
where we have collected the terms which do not depend on $\theta$ in $C$. By setting the derivative with respect to $\theta$ to zero, we obtain a criterion for stationarity
\begin{align}
\hspace{-0.2cm}
\boxed{
 \int\limits_{ y }\!\!\! \left(  \int\limits_{ x }\!\! \frac { \partial f(x,y,\theta ) }{ \partial \theta } q(x|y,\theta ) \!\! \right)\!\! p(y)\!=\!\!\int\limits_{ y,x }\! \frac { \partial f(x,y,\theta ) }{ \partial \theta } p(x,y)  . \! 
 }\label{eq:extrema} 
 \end{align}
 If we choose $f(\cdot)$ such that the objective (\ref{eq:objective_final}) is convex in $\theta$, then \eqref{eq:extrema} is a sufficient condition for optimality.
 
 Before this system of equations can be solved, all the integrals have to be computed. The integral over $x$ on the left-hand side of (\ref{eq:extrema}) is an expectation with respect to the parametric approximation. Since the integrand depends on unknown parameters, this inner integral cannot be approximated numerically. Therefore, $f$ has to be chosen such that there is a closed form solution.
 
 In general, the outer integral over $y$ cannot be solved in closed form since $p(y)$ can have a very complex form, depending on the dynamical system. However, expectations with respect to $p(y)$ can be efficiently approximated numerically, as discussed above. Numeric integration is possible only if the integrand depends on no other variable than the ones we integrate out.  Therefore, we require $f$ to be such that, after analytically solving the inner integral over $x$, all the dependences on $\theta$ can be moved outside of the integral over $y$.
 
 On the right-hand side of (\ref{eq:extrema}), we evaluate an expectation with respect to $p(x,y)$. Again, it is not possible for general dynamical systems to find a closed form solution, but numerical expectations with respect to $p(x,y)$ can be computed efficiently.  To allow for numerical integration, $f$ must be such that
all the dependences on $\theta$ can be moved outside of the integral over $x$ and $y$.
 
 Finally, after computing the integrals, we have to solve the system of equations (\ref{eq:extrema}) in order to find the optimal $\theta$.  Therefore, $f(\cdot)$ should be such that this solution can be found efficiently.
 
 It is not clear how the most general $q(x|y,\theta )$ complying with the above desiderata can be found. Nevertheless, this discussion can guide the search for more general belief representations than the affine Gaussian, which leave the efficiency of the GF intact.  The following section provides an example.

\section{The Feature Gaussian Filter}\label{sect:feature_gaussian_filter}
We propose to generalize the affine Gaussian approximate posterior of the GF by allowing for nonlinear features $\phi(y)$ of the measurement. More formally, we choose 
$f$ in (\ref{eq:approximate_distribution}) as
\begin{align}
f(x,y,\Gamma ,\Sigma )=-\frac { 1 }{ 2 } (x-\Gamma \phi (y))^{ T }\Sigma^{-1} (x-\Gamma \phi (y))\label{eq:conditional_gaussian_filter}
\end{align}
with parameters $\theta = (\Gamma, \Sigma)$ and $\phi$ an arbitrary feature function.
This leads to an approximate distribution (\ref{eq:approximate_distribution}), which is Gaussian in $x$ but can have nonlinear dependences on $y$,
\begin{align}
 q(x|y,\Gamma ,\Sigma )=\mathcal{N}(x|\Gamma \phi (y),\Sigma ) . \label{eq:approximate_distribution_cgf}
\end{align}
In the following, we show that because this approximation complies with the desiderata from the previous section, the parameters can be optimized efficiently. We refer to the resulting filtering algorithm as the Feature Gaussian Filter (FGF).
Finally, we show that the FGF is essentially equivalent to the standard GF using a virtual measurement, obtained by mapping the actual measurement through a nonlinear function.

\subsection{Finding $\Gamma$}
The derivative with respect to $\Gamma$ is
\begin{align}
 \frac { \partial f(x,y,\Gamma ,\Sigma) }{ \partial \Gamma  } =\Sigma^{-1} (x-\Gamma \phi (y))\phi (y)^{ T }
\end{align}
and the corresponding analytic integral can readily be solved since the approximate distribution is Gaussian in $x$:
\begin{align}
 \int\limits_{ x } \frac { \partial f(x,y,\Gamma ,\Sigma ) }{ \partial \Gamma  } q(x|y,\Gamma ,\Sigma )=0 .
\end{align}
Inserting these results into (\ref{eq:extrema}), we can solve for $\Gamma$
\begin{align}
 \boxed{
 \Gamma =E[x\phi (y)^{ T }]E[\phi (y)\phi (y)^{ T }]^{ -1 } .
 }\label{eq:gamma}
\end{align}

\subsection{Finding $\Sigma$}
The matrix $\Sigma$ is constrained to be positive definite, such that the approximate distribution (\ref{eq:approximate_distribution_cgf}) is Gaussian. As it turns out, the unconstrained optimization yields a positive definite matrix. Thus, there is no need to take this constraint into account explicitly.

The derivative with respect to $\Sigma^{-1}$ is
\begin{align}
 \frac { \partial f(x,y,\Gamma ,\Sigma) }{ \partial \Sigma^{-1}  } =-\frac { 1 }{ 2 } (x-\Gamma \phi (y))(x-\Gamma \phi (y))^{ T }
\end{align}
and the corresponding analytic integral in $x$ is 
\begin{align}
\int\limits_{ x } \frac { \partial f(x,y,\Gamma ,\Sigma) }{ \partial \Sigma^{-1}  } q(x|y,\Gamma ,\Sigma )=-\frac { 1 }{ 2 } \Sigma .
\end{align}
Inserting these results into (\ref{eq:extrema}),  we can solve for $\Sigma$
\begin{align}
 \boxed
 {
 \Sigma =E[(x-\Gamma \phi (y))(x-\Gamma \phi (y))^{ T }] .
 }\label{eq:omega}
\end{align}

\subsection{Connection to the Gaussian Filter}
In the following we show that for a feature $\phi(y)=(c,\varphi(y)^T)^T$, which contains a constant $c\neq 0$ and an arbitrary sub-feature $\varphi$, the FGF is equivalent to the GF using $\hat y =\varphi(y)$ as the measurement.
Inserting  $\phi(y)=(c,\hat y^T)^T$ into (\ref{eq:gamma}), we obtain
\begin{align}
\Gamma =\begin{pmatrix} \frac{\mu _{ x }-\Sigma _{ x\hat{y} }\Sigma _{ \hat{y}\hat{y} }^{ -1 }\mu _{ \hat{y} }}{c} & \Sigma _{ x\hat{y} }\Sigma _{ \hat{y}\hat{y} }^{ -1 } \end{pmatrix}
\end{align}
with the parameters $\mu_{(\cdot)}$ and $\Sigma_{(\cdot)}$ as defined in (\ref{eq:gf_expectations}). The mean of the approximate posterior is
\begin{align}
 \Gamma \phi (y)=\mu _{ x }+\Sigma _{ x\hat{y} }\Sigma _{ \hat{y}\hat{y} }^{ -1 }(\hat{y}-\mu _{ \hat{y} }) .
\end{align}
Inserting this result into (\ref{eq:omega}), we obtain the covariance
\begin{align}
 \Sigma=\Sigma _{ xx }-\Sigma _{ x\hat{y} }\Sigma _{ \hat{y}\hat{y} }^{ -1 }\Sigma _{ x\hat{y} }^{ T }.
\end{align}
Clearly, these equations correspond to the GF equations (\ref{eq:gaussian_filter_posterior}).  This means that, if the feature vector $\phi(y)$ contains a constant, the FGF is equivalent to the GF using the virtual measurement $\hat{y} = \varphi(y)$ instead of $y$. In particular, with a feature $\phi(y)=(1,y^T)^T$, we retrieve the standard GF.

Applying nonlinear transformations to the physical sensor measurements before feeding them into a GF is not uncommon in robotics and other applications (see \cite{radar_tracking,attitude_estimation,humanoid_robot} for example).  The formal analysis herein provides insight into the effect of such nonlinear transformations and reveals why they are beneficial. Namely, they allow for a better fit of the conditional distribution.  While these transformations 
are often motivated from physical insight or introduced heuristically, we provide a different interpretation of $\phi$ as a means of improving the fit of the posterior by allowing for more expressive nonlinear features.  This shall be highlighted in the examples in Section \ref{sect:PerformanceAnalysis}, where we use monomials of increasing order as generic features.


\subsection{Feature Selection}
The above analysis shows
that adding nonlinear features gives the approximate distribution more flexibility to fit the exact distribution. Overfitting is not possible since we are minimizing the KL-divergence to the exact distribution. It therefore makes sense to use as many features as the computational speed requirements allow.

Ideally, one would choose a feature which maps the measurement to a representation which relates to the state linearly. If this is not possible, then generic features such as monomials can be used.

\subsection{Computational Complexity}
The only cause of a difference in computational complexity between the standard GF and the FGF is the difference in the dimension of the measurement $y$ and the feature $\phi(y)$.
This means that the feature dimension has to be chosen such that the required computational speed is attained. The feature dimension can even be lower than the dimension of the actual measurement if the standard GF is too slow.

\section{Analysis and Simulation of the Feature Gaussian Filter}\label{sect:PerformanceAnalysis}
As the previous analysis suggests, it is beneficial to augment the measurement with nonlinear features since this gives the approximation more flexibility to fit the exact distribution, i.e. to achieve a lower KL-divergence \eqref{eq:objective_final}. In this section, we illustrate this effect in more detail for two dynamical systems. 

\subsection{Estimation of Sensor Noise Magnitude}
The measurement process \eqref{eq:functional_observation_model} of a dynamical system can often be represented by a nonlinear observation model with additive noise
\begin{align}
h(x,M,w)=\tilde{h}(x)+Mw \label{eq:additive_noise_observation_model}
\end{align}
where $\tilde{h}$ is a nonlinear function of the system state, and the matrix $M$ determines the magnitude of the sensor noise (recall that $w$ is Gaussian with zero mean and unit variance). 
Often, the sensor accuracy (i.e. the matrix $M$) is not precisely known, or it may be time varying due to changing sensor properties and environmental conditions. It is then desirable to estimate the noise matrix $M$ alongside the state $x$. In the following, we show that this is not possible with the standard GF, but can be achieved with the FGF. 

We define an augmented state $\hat x := (x; m)$, where m is a column vector containing all the elements of the noise matrix $M$. The observation model in distributional form is $p(y|\hat x)=p(y|x,m)=\mathcal{N}(y|\tilde{h}(x),MM^T)$. The state $x$ and the parameters $m$ stem from independent processes, and we therefore have $p(\hat x)=p(x)p(m)$. Let us now apply the standard GF to this problem by computing the parameters in (\ref{eq:gf_expectations}). In particular, we compute the covariance between the augmented state and the measurement
\begin{align}
\Sigma_{\hat x y}&=\int\limits_{x,m, y}\!\!\!\begin{pmatrix} x-\mu_x\\m- \mu_m\end{pmatrix} (y-\mu_{y})^Tp( y|x,m)p(x)p(m). \label{eq:augmented_covariance}
\end{align}
The integral over $y$ can be solved easily since $p(y|x,m)$ is Gaussian,
\begin{align}
\Sigma_{\hat x y}&=\int\limits_{x,m}\begin{pmatrix} x-\mu_x\\m- \mu_m\end{pmatrix}( \tilde{h}(x)-\mu_{y})^Tp(x)p(m) .
\end{align}
Interestingly, the second factor does not depend on $m$.  Therefore, the integral over $m$ is solved easily and yields
\begin{align} \Sigma _{ \hat { x } y } & =\int _{ x } \begin{pmatrix} x-\mu _{ x } \\ \mu _{ m }-\mu _{ m } \end{pmatrix}(\tilde{h}(x)-\mu _{ y })^{ T }p(x)=\begin{pmatrix} \Sigma _{ xy } \\ 0 \end{pmatrix}. \end{align}
As a result, there is no linear correlation between the measurement $y$ and the parameters $m$. Inserting this result into  (\ref{eq:gaussian_filter_posterior}) shows that the innovation corresponding to $m$ is zero. The corresponding part of the covariance matrix does not change either.  The measurement has hence no effect on the estimate of $m$. It will behave as if no observation had been made. This illustrates the failure of the GF to capture certain dependences in nonlinear dynamical systems.

In contrast, if a nonlinear feature in the measurement $y$ is used, the integral over $y$ in (\ref{eq:augmented_covariance}) will not yield $\tilde{h}(x)$, but instead some function depending on both $x$ and $m$. This dependence allows the FGF to infer the desired parameters.

\subsubsection*{Numerical example}
For the purpose of illustrating the theoretical argument above, we use a small toy example. We consider a single sensor, where all quantities in (\ref{eq:additive_noise_observation_model}), including the standard deviation $M$, are scalars.  Since we are only interested in the estimate of $M$, we choose $\tilde h(x)=0$. The observation model (\ref{eq:additive_noise_observation_model}) simplifies to
\begin{align}
 h(M_2,w_2)=M_2 w_2 .
\end{align}
Note that we have reintroduced time indices.
Picking a simple process model and an initial distribution
\begin{align}
 g(M_{ 1 },v_{ 2 })&=M_{ 1 }+0.1v_{ 2 }\\ 
 p(M_{ 1 })&=\mathcal{N}(M_{ 1 }|5,1)
\end{align}
the dynamical system \eqref{eq:functional_process_model}, \eqref{eq:functional_observation_model} is fully defined.
This example captures the fundamental properties of the FGF as pertaining to the estimation of sensor noise intensity $M$.  The same qualitative effects hold for multivariate systems (\ref{eq:additive_noise_observation_model}) for the reasons stated above.

\begin{figure}[tb]
  \centering
  	\includegraphics[width=1.0\linewidth]{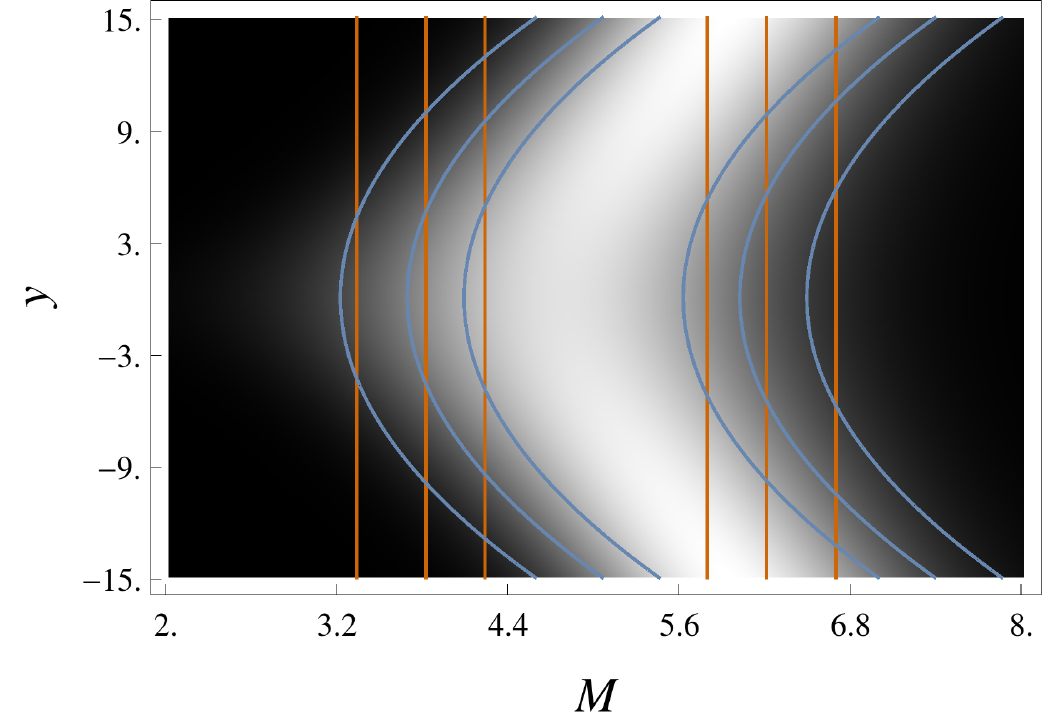}
  	\vspace{-0.2cm}
 	 \caption{Estimation of sensor noise magnitude: Density plot of the true conditional distribution $p(M_2|y_2)$ with overlaid contour lines of the approximate conditional distribution $q(M_2|y_2)$ of the GF in orange and of the FGF in blue.}
  	\label{fig:noise_approximate_vs_exact}
\end{figure}

In Figure \ref{fig:noise_approximate_vs_exact}, we plot the exact conditional distribution $p(M_2|y_2)$ implied by this system in grayscale. This distribution was computed numerically for the purpose of comparison. It would, of course, be too expensive to use in a filtering algorithm. 
The overlaid orange contour lines show the approximate conditional distribution $q(M_2|y_2)$ obtained with the standard GF.
No matter what measurement $y_2$ is obtained, the posterior $q(M_2|y_2)$ is the same. The GF does not react to the measurements at all.

The true conditional distribution $p(M_2|y_2)$ depends on $y_2$, which means that the measurement does in fact contain information about the state $M_2$. However, the approximation $q(M_2|y_2)$ made by the GF is not expressive enough to capture this information, which results in a very poor fit to $p(M_2|y_2)$.

The standard GF is the special case of the FGF with the feature $\phi (y)=(1, y)^T$. Let us take the obvious next step and add a quadratic term to the feature $\phi (y)=(1, y, y^2)^T$. The resulting approximation is represented by the blue contour lines in Figure \ref{fig:noise_approximate_vs_exact}. 
Clearly, $q(x_2|y_2)$ now depends on the measurement $y_2$, which allows the FGF to exploit the 
information  about the state $x_2$ contained in the measurement. The approximation $q(x_2|y_2)$ of the FGF has a more flexible form, which allows for a better fit of the true posterior.

\begin{figure}[tb]
  \centering
  	\includegraphics[width=1.0\linewidth]{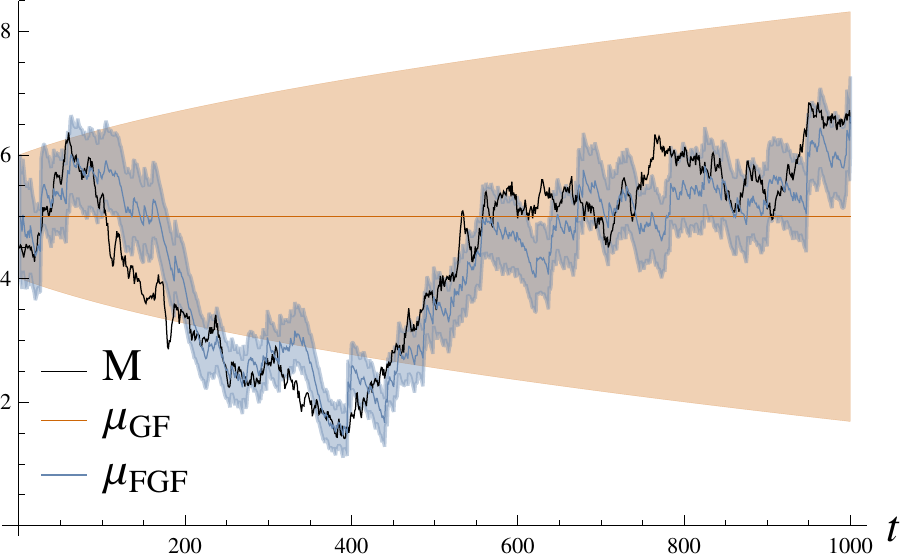}
 	 \caption{Estimation of sensor noise magnitude: The simulated noise parameter $M$ is shown in black, together with the mean and standard deviation of the estimates obtained with the GF (orange) and the FGF (blue).}
  	\label{fig:noise_filtering}
\end{figure}

To analyse actual filtering performance, we simulate the dynamical system and the two filters for 1000 time steps.  The results are shown in Figure \ref{fig:noise_filtering}. 
As expected, the standard GF does not react in any way to the incoming measurements. 
The FGF, on the other hand, is capable of inferring the state $M$ from the measurement $y$, as suggested by the theoretical analysis above.

\subsection{Nonlinear Observation Model}
In this section, we investigate how the theoretical benefit of adding nonlinear features translates into improved filtering performance for systems with nonlinear observation models.  To clearly illustrate the difference of GF and FGF, we choose a simple system with a strong nonlinearity (step function).  
Given the theoretical analysis herein, it is to be expected that the insights gained from this artificial example extend to more realistic nonlinear problems in robotics and other applications.

The process model, the observation model, and the initial state distribution are given by
\begin{align}
 g(x_{ 1 },v_{ 2 })&=x_{ 1 }+v_{ 2 }\\ 
 h(x_{ 2 },w_{ 2 })&=x_{ 2 }+w_{ 2 }+50H(x_{ 2 })\\ 
 p(x_{ 1 })&=\mathcal{N}(x_{ 1 }|0,5)
\end{align}
where $H(\cdot)$ is the Heaviside step function.

\begin{figure}[tb]
  \centering
  	\includegraphics[width=1.0\linewidth]{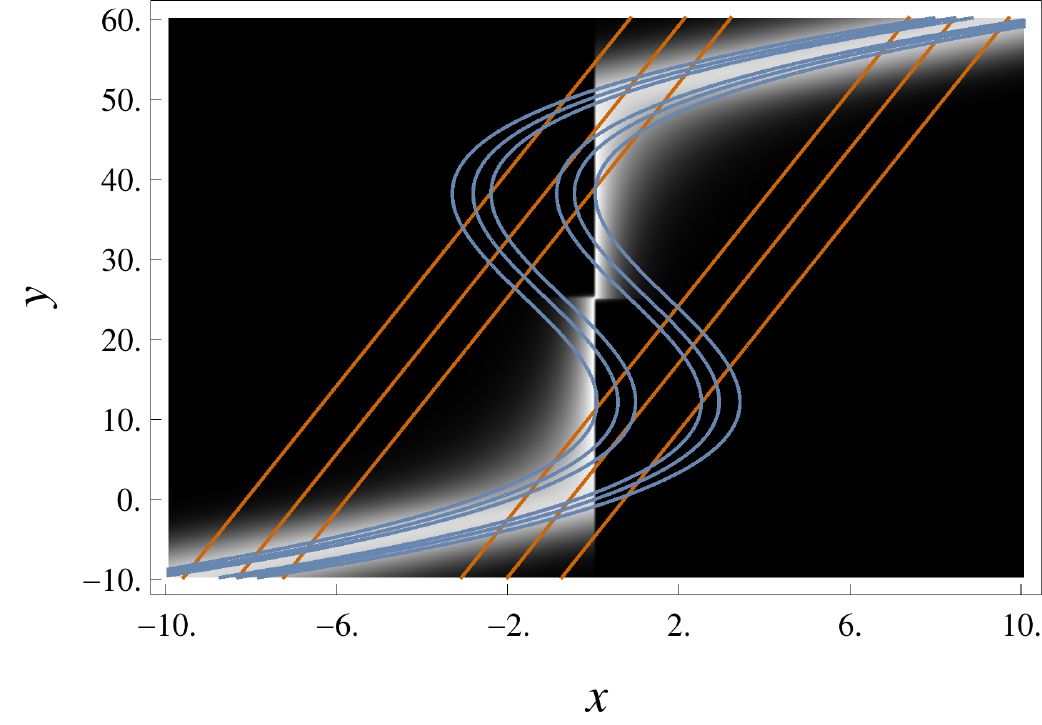}
  	  	\vspace{-0.2cm}
 	 \caption{ Nonlinear observation model: Density plot of the true conditional distribution $p(x_2|y_2)$ with overlaid contour lines of the approximate conditional distribution $q(x_2|y_2)$ of the GF in orange and of the FGF in blue.}
  	\label{fig:nonlinear_approximate_vs_exact}
\end{figure}

In Figure \ref{fig:nonlinear_approximate_vs_exact}, we plot the true conditional density $p(x_2|y_2)$ with overlaid orange contour lines of the approximate conditional distribution $q(x_2|y_2)$ obtained using the standard GF. 
The contour lines reflect the estimator structure of the GF in (\ref{eq:gaussian_filter_posterior}). The mean of the approximate density $q(x_2|y_2)$ is an affine function of the measurement $y_2$. For nonlinear observation models, this coarse approximation can lead to loss of valuable information contained in the measurement $y_2$. 

The approximate density $q(x_2|y_2)$ obtained using a feature $\phi (y)=(1, y, y^2, y^3)^T$, which is represented by the blue contour lines in Figure \ref{fig:nonlinear_approximate_vs_exact}, fits the true posterior much better.
This  illustrates that nonlinear features allow for approximate posteriors with much more elaborate dependences on $y$.

\begin{figure}[tb]
  \centering
  	\includegraphics[width=1.0\linewidth]{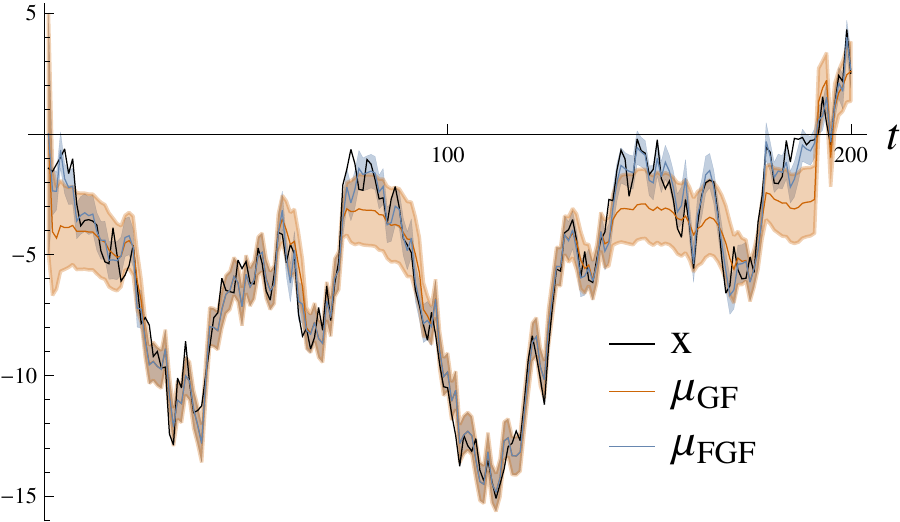}
 	 \caption{Nonlinear observation model: We plot the simulated state $x$ (black) and the means and standard deviations of the estimates obtained with the GF (orange) and the FGF (blue).}
  	\label{fig:nonlinear_filtering}
\end{figure}

Figure \ref{fig:nonlinear_filtering} shows how this difference translates to filtering performance.
When $x$ is far away from zero, the nonlinearity has no effect: the system behaves like a linear system. The density plot in this regime would be centered at a linear part of the distribution, and both filters would achieve a perfect fit. Both the standard GF and the FGF are therefore optimal in that case. When the state is close to zero, however, the advantage of the FGF becomes apparent. Its tracking performance is good even when the state is close to the nonlinearity of the observation model, due to more flexibility in $y_2$ of the posterior approximation $q(x_2|y_2)$.

\section{Conclusion}
We showed that the GF can be understood as an optimal approximation to the exact distribution, subject to the constraint that the form of the belief $q(x|y)$ be Gaussian in $x$ and affine in $y$. Theoretical analysis and simulations showed that this form can be too restrictive to accurately represent the belief in nonlinear systems. We discussed how this constraint can be relaxed while maintaining the efficiency of the GF. This analysis
served as a basis for potential generalizations of the GF.

We proposed one such generalization, the Feature Gaussian Filter (FGF). 
The name is motivated by the fact that the FGF is
equivalent to a GF that uses a virtual measurement, or feature, which is obtained by applying a nonlinear function to the actual measurement.
We showed both theoretically and in simulation that using nonlinear features can significantly improve the performance of the GF. For instance, the practically relevant problem of estimating the sensor noise magnitude alongside the state cannot be tackled by the standard GF because the expressive power of its belief is too limited. We showed that this issue can be resolved by the FGF.


The results obtained in the simulation examples herein are promising and suggest that the FGF may yield superior filtering performance for nonlinear problems in robotics and other applications. Analysing the performance of the FGF in a more realistic, high dimensional scenario remains future work. 

Whether it is possible to find an approximate posterior of a more general form than in the FGF, while complying with the requirements derived in Section \ref{sect:generalization}, is another open question.

\pagebreak



\bibliographystyle{plainnat}
\bibliography{../references}

\begin{thebibliography}{21}
\providecommand{\natexlab}[1]{#1}
\providecommand{\url}[1]{\texttt{#1}}
\expandafter\ifx\csname urlstyle\endcsname\relax
  \providecommand{\doi}[1]{doi: #1}\else
  \providecommand{\doi}{doi: \begingroup \urlstyle{rm}\Url}\fi

\bibitem[Barber(2012)]{barber}
D.~Barber.
\newblock \emph{{Bayesian Reasoning and Machine Learning}}.
\newblock {Cambridge University Press}, 2012.

\bibitem[Bene{\v{s}}(1981)]{benes}
V.E. Bene{\v{s}}.
\newblock Exact finite-dimensional filters for certain diffusions with
  nonlinear drift.
\newblock \emph{Stochastics: An International Journal of Probability and
  Stochastic Processes}, 1981.

\bibitem[Bickel et~al.(2008)Bickel, Li, and Bengtsson]{bickel}
P.~Bickel, B.~Li, and T.~Bengtsson.
\newblock Sharp failure rates for the bootstrap particle filter in high
  dimensions.
\newblock In \emph{IMS Collections: Pushing the Limits of Contemporary
  Statistics}. 2008.

\bibitem[Cappe et~al.(2007)Cappe, Godsill, and Moulines]{cappe_overview}
O.~Cappe, S.J. Godsill, and E.~Moulines.
\newblock An overview of existing methods and recent advances in sequential
  {M}onte {C}arlo.
\newblock \emph{Proceedings of the IEEE}, 2007.

\bibitem[Daum(1986)]{daum}
F.E. Daum.
\newblock Exact finite-dimensional nonlinear filters.
\newblock \emph{IEEE Transactions on Automatic Control}, 1986.

\bibitem[Daum and Fitzgerald(1983)]{radar_tracking}
F.E. Daum and R.J. Fitzgerald.
\newblock Decoupled {K}alman filters for phased array radar tracking.
\newblock \emph{IEEE Transactions on Automatic Control}, 1983.

\bibitem[Gordon et~al.(1993)Gordon, Salmond, and Smith]{gordon}
N.J. Gordon, D.J. Salmond, and A.F.M. Smith.
\newblock Novel approach to nonlinear/non-{G}aussian bayesian state estimation.
\newblock \emph{IEE Proceedings F (Radar and Signal Processing)}, 1993.

\bibitem[Ito and Xiong(2000)]{gf}
K.~Ito and K.~Xiong.
\newblock Gaussian filters for nonlinear filtering problems.
\newblock \emph{IEEE Transactions on Automatic Control}, 2000.

\bibitem[{Julier} and {Uhlmann}(1997)]{ukf}
S.J. {Julier} and J.K. {Uhlmann}.
\newblock A new extension of the {K}alman filter to nonlinear systems.
\newblock In \emph{Proceedings of AeroSense: The 11th Int. Symp. on
  Aerospace/Defense Sensing, Simulations and Controls}, 1997.

\bibitem[Kalman(1960)]{kalman1960new}
R.E. Kalman.
\newblock A new approach to linear filtering and prediction problems.
\newblock \emph{Journal of Basic Engineering}, 1960.

\bibitem[Kushner(1967)]{earlyKushner}
H.J. Kushner.
\newblock Approximations to optimal nonlinear filters.
\newblock \emph{IEEE Transactions on Automatic Control}, 1967.

\bibitem[Kushner and Budhiraja(2000)]{kushner}
H.J. Kushner and A.S. Budhiraja.
\newblock A nonlinear filtering algorithm based on an approximation of the
  conditional distribution.
\newblock \emph{IEEE Transaction on Automatic Control}, 2000.

\bibitem[Li et~al.(2005)Li, Bengtsson, and Bickel]{li}
B.~Li, T.~Bengtsson, and P.~Bickel.
\newblock Curse-of-dimensionality revisited: Collapse of importance sampling in
  very large scale systems.
\newblock Technical report, Department of Statistics, UC-Berkeley, 2005.

\bibitem[Morelande and Garcia-Fernandez(2013)]{garcia}
M.R. Morelande and A.F. Garcia-Fernandez.
\newblock Analysis of {K}alman filter approximations for nonlinear
  measurements.
\newblock \emph{IEEE Transactions on Signal Processing}, 2013.

\bibitem[N{\o}rgaard et~al.(2000)N{\o}rgaard, Poulsen, and Ravn]{ddf}
M.~N{\o}rgaard, N.K. Poulsen, and O.~Ravn.
\newblock New developments in state estimation for nonlinear systems.
\newblock \emph{Automatica}, 2000.

\bibitem[Owen(2013)]{owen}
A.B. Owen.
\newblock Monte carlo theory, methods and examples (book draft), 2013.
\newblock URL \url{http://statweb.stanford.edu/~owen/mc/}.

\bibitem[Rotella et~al.(2014)Rotella, Bloesch, Righetti, and
  Schaal]{humanoid_robot}
N.~Rotella, M.~Bloesch, L.~Righetti, and S.~Schaal.
\newblock State estimation for a humanoid robot.
\newblock In \emph{IEEE/RSJ International Conference on Intelligent Robots and
  Systems}, 2014.

\bibitem[S{\"a}rkk{\"a}(2013)]{sarkka}
S.~S{\"a}rkk{\"a}.
\newblock \emph{Bayesian filtering and smoothing}.
\newblock Cambridge University Press, 2013.

\bibitem[Sorenson(1960)]{ekf}
H.W. Sorenson.
\newblock \emph{Kalman Filtering: Theory and Application}.
\newblock IEEE Press selected reprint series. IEEE Press, 1960.

\bibitem[Vaganay et~al.(1993)Vaganay, Aldon, and Fournier]{attitude_estimation}
J.~Vaganay, M.J. Aldon, and A.~Fournier.
\newblock Mobile robot attitude estimation by fusion of inertial data.
\newblock In \emph{IEEE International Conference on Robotics and Automation},
  1993.

\bibitem[Wu et~al.(2006)Wu, Hu, Wu, and Hu]{wu}
Y.~Wu, D.~Hu, M.~Wu, and X.~Hu.
\newblock A numerical-integration perspective on {G}aussian filters.
\newblock \emph{IEEE Transactions on Signal Processing}, 2006.

\end{thebibliography}

\end{document}